\documentclass[runningheads]{llncs}

\usepackage{eccv}

\usepackage{eccvabbrv}

\usepackage{graphicx}
\usepackage{booktabs}

\usepackage[accsupp]{axessibility}  %

\usepackage{hyperref}

\usepackage{orcidlink}

\usepackage{graphicx}
\usepackage{amsmath}
\usepackage{amssymb}
\usepackage{booktabs}
\usepackage{ragged2e}

\usepackage{color}
\usepackage{amssymb}

\usepackage{lipsum}
\usepackage{xcolor}
\usepackage{tabularx}
\usepackage{multirow}
\usepackage{enumitem}
\usepackage{bbm}
\usepackage{wrapfig}
\usepackage{setspace}
\usepackage{footnote}

\usepackage{colortbl} %
\usepackage{pifont}  %
\usepackage{mathtools}  %
\usepackage[font=small]{caption}  %
\usepackage{blindtext}  %
\usepackage{stmaryrd} %
\usepackage[accsupp]{axessibility}  %
\usepackage{mathtools}
\usepackage{subfiles}
\usepackage{afterpage}

\usepackage{algorithm}
\usepackage{algpseudocode}
\usepackage{adjustbox}

\usepackage{anyfontsize}
\usepackage{afterpage}
\usepackage{xcolor,pifont}

\definecolor{amethyst}{rgb}{0.6, 0.4, 0.8}

\definecolor{grey}{rgb}{0.9, 0.9, 0.9}
\newcommand{\ccol}{\cellcolor{grey}}
\newcommand*\colourcheck[1]{%
  \expandafter\newcommand\csname #1check\endcsname{\textcolor{#1}{\ding{51}}}%
}

\colourcheck{blue}
\colourcheck{green}
\colourcheck{red}
\newcommand{\cmark}{\ding{51}}%
\newcommand{\xmark}{\ding{55}}%
\newcommand{\expnum}[2]{{#1}\mathrm{e}{-#2}}
\newcommand{\sftype}[1]{{\textsf{\small #1}}}

\def\ie{\emph{i.e.}}
\def\eg{\emph{e.g.}}
\def\etal{\emph{et al.}}

\definecolor{bright_red}{rgb}{0.97, 0.9, 0.9}
\definecolor{blk}{rgb}{0, 0, 0}
\definecolor{mgt}{rgb}{0.8, 0.1, 0.8}
\definecolor{darkblue}{rgb}{0.2, 0.2, 0.8}
\definecolor{lblue}{rgb}{0.2, 0.2, 1.0}
\definecolor{orange}{rgb}{1.0, 0.5, 0.2}
\definecolor{goldenrod}{rgb}{0.85, 0.65, 0.33}
\definecolor{darkred}{rgb}{0.75, 0.0, 0.0}
\definecolor{grn}{rgb}{0.2, 0.6, 0.2}

\begin{document}

\title{Learning Sample-wise Rank-aware Interpolation Weights for Composed Visual Data Retrieval}

\titlerunning{Learning Sample-wise Rank-Aware Interpolation for Composed Retrieval}

\author{Boseung~Jeong\inst{1}\orcidlink{0000-0001-9382-3396}
    \and Taegyu~Park\inst{2}\orcidlink{0009-0006-6387-644X}
    \and Donghyeon~Kwon\inst{2}\orcidlink{0009-0006-2791-1471}
    \and \\ Hyunsouk~Cho\inst{3}\orcidlink{0000-0002-9134-1921}
    \and Suha~Kwak\inst{2,4}\orcidlink{0000-0002-4567-9091}}

\authorrunning{B.~Jeong et al.}

\institute{\textsuperscript{1}AI Center, Samsung Electronics, Republic of Korea
    \\ \textsuperscript{2}Dept. of CSE, POSTECH, Republic of Korea
    \\ \textsuperscript{3}Dept. of AI, Ajou University, Republic of Korea
    \\ \textsuperscript{4}Graduate School of AI, POSTECH, Republic of Korea
    \\ \email{\inst{1}bs0131.jeong@samsung.com,
        \inst{2}\{taegyu.park,kinux98,suha.kwak\}@postech.ac.kr,
        \\ \inst{3}hyunsouk@ajou.ac.kr}}

\maketitle

\begin{abstract}
    At the heart of composed visual data retrieval is the fusion of a reference visual input and a textual modification into a single query.
    While current state-of-the-art methods utilize multimodal large language models for this fusion, their complexity introduces prohibitive query-time latency, limiting their scalability.
    We instead revisit the efficacy of simple linear interpolation within an embedding space, and introduce SRAIN, the first framework that dynamically predicts query-specific interpolation weights.
    The key challenge lies in the fact that the quality of an interpolation weight should be measured by the interpolated embedding's discriminability from negatives as well as its proximity to true targets; this makes collecting and predicting optimal weights intractable.
    We overcome this bottleneck through two key innovations: batch-wise rank-aware weight estimation during training, and a compact memory bank that synthesizes hard negatives during inference.
    SRAIN achieves the best in composed video retrieval and
    matches the current state of the art in composed image retrieval, all while substantially reducing query-time latency compared to MLLM-based alternatives.
\end{abstract}

\section{Introduction}
\label{sec:intro}

\begin{figure}[!t]
    \centering
    \includegraphics[width=\linewidth]{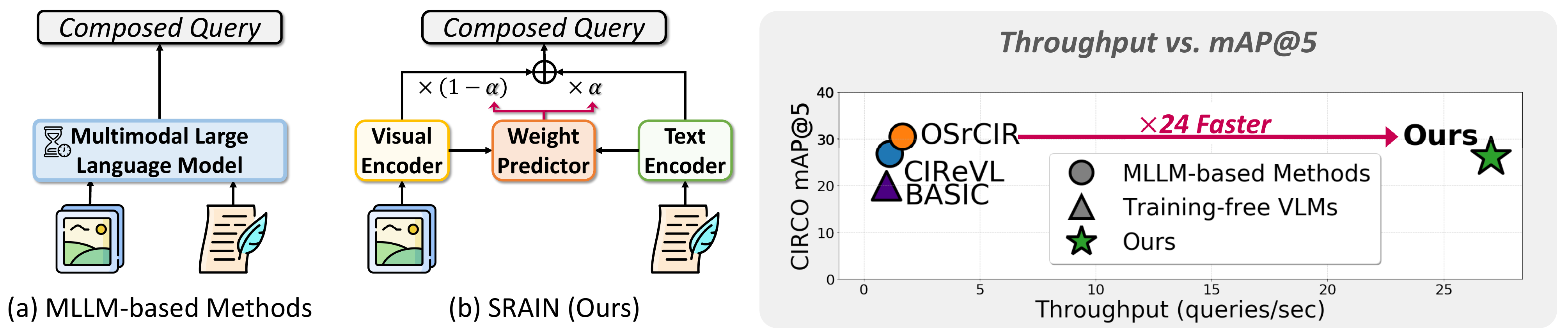}
    \caption{Comparison of SRAIN with MLLM-based methods. (a) MLLM-based methods process reference and modification text together through MLLMs to generate a composed query, introducing substantial computational overhead. (b) SRAIN predicts a sample-specific interpolation weight $\alpha$ and fuses reference and modification embeddings through linear interpolation, enabling efficient retrieval.
        The right plot shows that SRAIN achieved comparable performance (mAP@5 on CIRCO) while being $\times$24 faster in throughput compared to MLLM-based methods.
        Meanwhile, BASIC~\cite{basic}, a training-free method without MLLM, underperformed in throughput as well as retrieval accuracy since it requires numerous text encoder forward passes.
    }
    \label{fig:teaser}
\end{figure}

Composed image retrieval (CoIR)~\cite{vo2019cir} and composed video retrieval (CoVR)~\cite{webvid2024} formulate the retrieval task as identifying a target visual instance based on a composite query: a reference visual input and an accompanying textual modification.
This approach improves search specificity and expands practical applications in media editing, recommendation systems, and creative content search.
Also, its inherent query editing capability has driven growing attention across various applications where modifying existing media is preferred over initiating a completely new search.

Both of CoIR and CoVR demand a principled fusion of the reference and modification text into a single query that reflects the intended edit while preserving the salient content of the reference.
A large body of research has been devoted to the multimodal fusion in this context, \eg, via cross-attention between visual and textual embeddings~\cite{Wan2024cross, wen2023target, zhang2021heterogeneous, sun2024image, zhao2024neucore, xu2023multi, simple2025, thawakar2024composed}
or textual inversion that encodes visual data in the text embedding space~\cite{saito2023pic2word,tang2024context,lin2024fine}.
Currently, the best-performing approach leverages multimodal large language models (MLLMs)~\cite{llava, lin2024videollava, achiam2023gpt4, brown2020language}
to rewrite the paired inputs into a single natural language description, which is in turn used for text-based retrieval~\cite{cirevl, simple2025, yang2024ldre, hummel2024egocvr, thawakar2024composed}.
However, the use of MLLMs substantially increases query-time latency due to their architectural complexity, limiting the practical utility of this approach.

Meanwhile, it has been recently reported that fusing the reference and modification embeddings by linear interpolation on a unit hyper-sphere can yield effective composed queries without auxiliary fusion modules~\cite{jang2024slerp}.
This work manually seeks a single interpolation weight for each dataset, which is however highly sub-optimal since the impact of the modification text should be determined by the specific content of the reference and the intent of the composite query.

To overcome this drawback while retaining the efficiency of linear interpolation,
we introduce a novel framework, dubbed \textbf{S}ample-wise \textbf{R}ank-\textbf{A}ware \textbf{IN}terpolation for multimodal fusion (SRAIN), that for the first time learns to determine how strongly a textual modification should influence the reference in an instance-wise manner.
At inference,
a lightweight module predicts an interpolation weight for each composite query, and its reference and modification embeddings are linearly interpolated using the predicted weight.
The resulting fused embedding is compared with target embeddings through cosine similarity for retrieval.
This pipeline enables efficient inference without any dependence on external models such as MLLMs~(\cref{fig:teaser}).

\begin{figure}[!t]
    \centering
    \includegraphics[width=\linewidth]{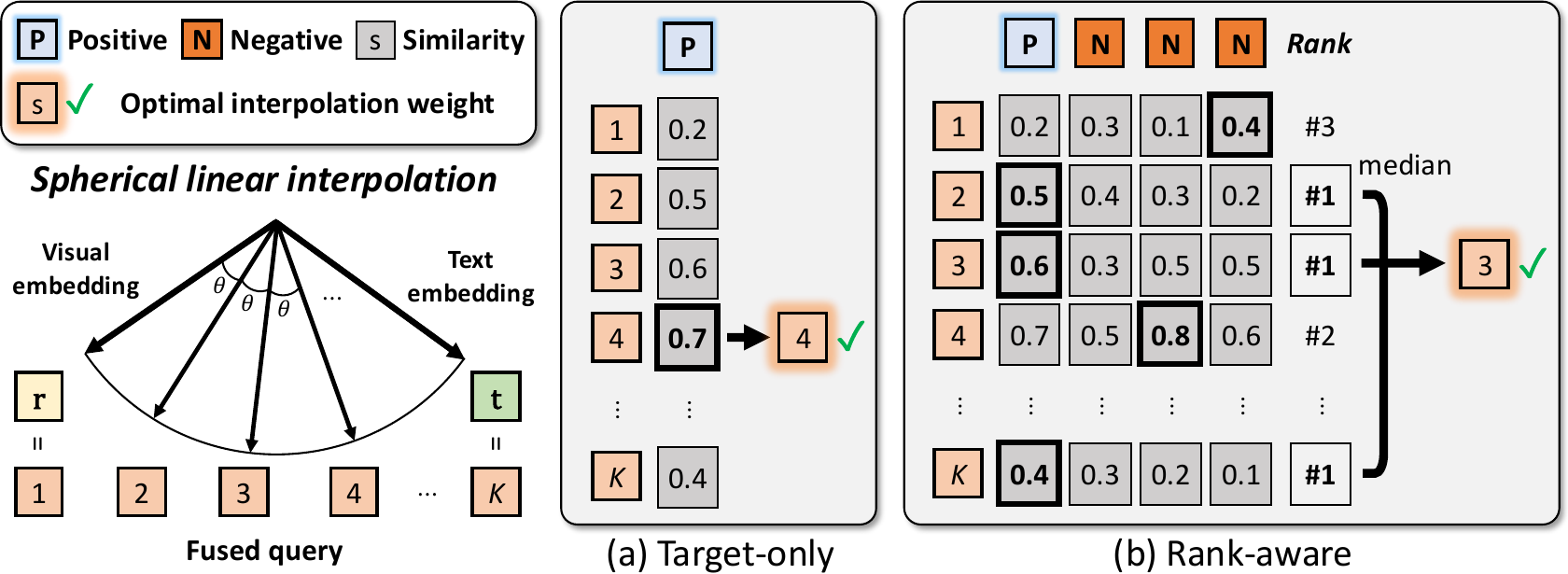}
    \caption{Rank-aware optimal weight estimation.
        Rather than determining the weight solely based on the similarity to the positive target (a), SRAIN selects the weight based on rank, where the positive consistently outranks nearby hard negatives (b).
    }
    \label{fig:ranking}
\end{figure}

However, training the interpolation weight predictor is challenging, primarily due to the lack of ground-truth supervision.
Identifying the optimal interpolation weight is computationally intractable; such a value cannot be determined solely by observing the composite query, but should be estimated by evaluating how effectively the interpolated embedding discriminates true targets from hard negatives.
Given the massive scale of the dataset, verifying this discriminative power against all positive and negative examples for every potential weight candidate is computationally intractable.
To tackle this problem, we propose a batch-wise rank-aware weight estimation strategy~(\cref{fig:ranking}). This strategy derives a near-optimal weight from ranking behavior within each mini-batch of a sufficiently large size, and offers the estimated weight as the ground-truth label.

\begin{figure*}[t!]
    \centering
    \includegraphics[width=1.0\linewidth]{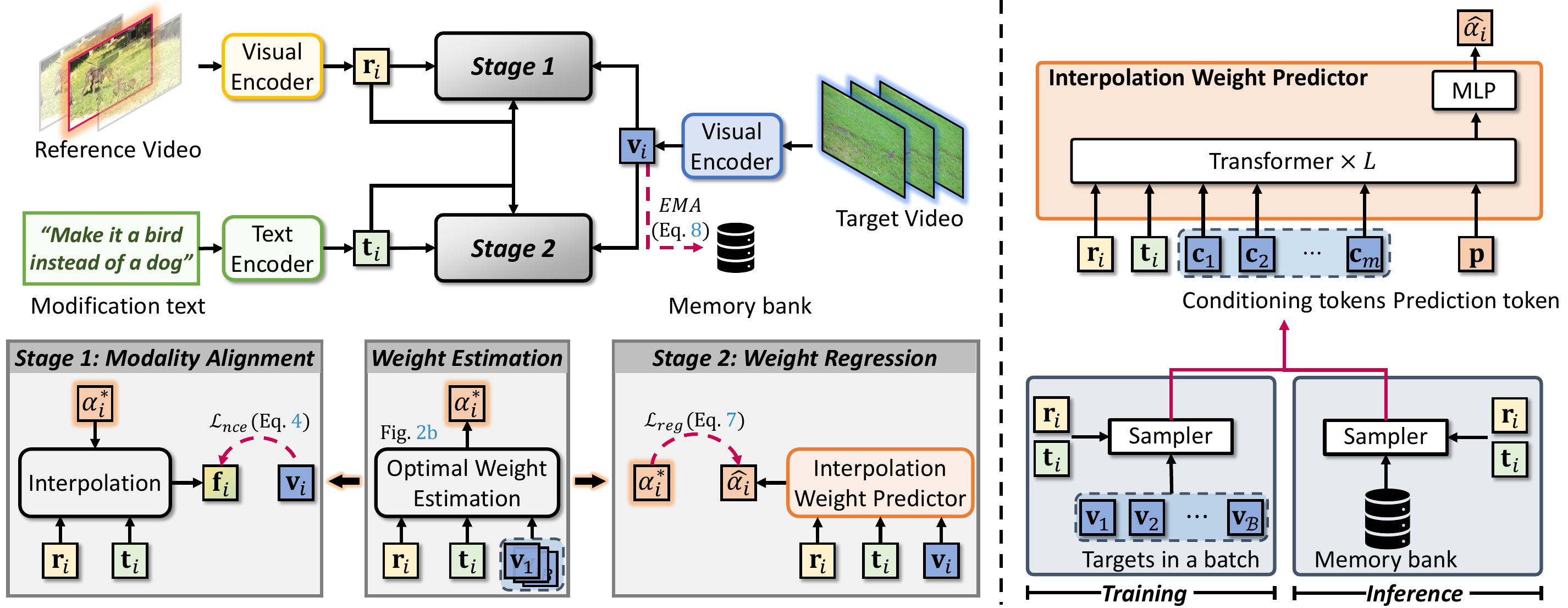}
    \caption{
        Overall architecture of SRAIN. The reference video, modification text, and target video are encoded using the pretrained
        encoders.
        For each batch, optimal interpolation weights are estimated in a rank-aware manner. The model is trained in two stages: In $Stage~1$,
        the encoders are fine-tuned to align composite query embeddings with target embeddings through hard-negative contrastive learning.
        In $Stage~2$,
        the encoders are frozen and a lightweight weight predictor is trained to regress $\alpha_i^*$. A memory bank is updated via exponential moving average~(EMA) during training to collect representative target prototypes. At inference, the memory bank supplies top-$n$ prototypes as auxiliary tokens, enabling consistent weight prediction.
    }
    \label{fig:overall}
    \vspace{-2ex}
\end{figure*}

Since the ground-truth weight depends on relative ranks with hard negatives in the batch, it is infeasible to infer the optimal weight solely from the reference and modification embeddings.
We mitigate this issue by introducing a conditioned weight prediction mechanism, which enables the weight predictor to reference hard negatives in both training and testing (\cref{fig:overall}).
During training, the weight predictor takes as inputs a positive target and hard negatives within the batch as well as the composite query embeddings.
At inference time, it samples synthetic negative embeddings from a compact memory bank that stores representative target embeddings of training data.

Building on these designs, we present a two-stage training strategy.
In the first stage, the visual and text encoders are fine-tuned by aligning the interpolated embedding with its corresponding target embedding while pushing apart negatives through a hard-negative contrastive loss~\cite{radenovic2023hnnce}.
In the second stage, the encoders are frozen, and only the weight predictor is trained to regress the ground-truth interpolation weights.
This separation is essential since the joint optimization of the encoders and the weight predictor causes the ground-truth interpolation weights to fluctuate as the encoder representation changes, which results in inconsistent supervision and unstable training.

SRAIN achieved the state of the art on the CoVR benchmark~\cite{webvid2024}, and matched the best model on the CoIR benchmarks~\cite{CIRCO, FIQ, CIRR}, all without relying on external models such as LLMs.
More importantly, it substantially reduces query-time latency compared to recent methods using MLLMs across both tasks.
The main contribution of this paper is three-fold:
\begin{itemize}[leftmargin=*, topsep=1mm]
        \item We propose SRAIN, the first framework to dynamically predict a per-sample interpolation weight to fuse the reference and modification embeddings while preserving computational efficiency.
    \item We address the lack of ground-truth supervision and the dependence of optimal weights on hard negatives by introducing a rank-aware weight estimation procedure that derives near-optimal weights from batch-wise ranking behavior, and a conditioned weight prediction mechanism that explicitly conditions on hard negatives. %
    \item SRAIN achieves the state of the art on the CoVR benchmark, while matching the best model on the CoIR benchmarks without reliance on external models, reducing query-time latency substantially.
\end{itemize}

\section{Related Work}
\label{sec:related_work}
\subsection{Composed Image Retrieval}
Composed Image Retrieval (CoIR)~\cite{vo2019cir} aims to retrieve a target image given a reference image and a textual modification that specifies the desired change.
This task extends traditional image-text retrieval~\cite{sun2021lightningdot} to a compositional setting and requires both modality alignment and multimodal fusion.
Prior methods~\cite{clip4cir, CIRR,zhang2021heterogeneous,xu2023multi,wen2023target,Wan2024cross,sun2024image,zhao2024neucore,zhang2024magiclens} have adopted the cross-attention mechanism to integrate visual and textual cues.
Another line of studies~\cite{cirevl,tang2025reason} has leveraged MLLMs to compose visual and textual semantics without task-specific training.
These methods have demonstrated strong generalization capability without training, but they depend on computationally intensive inference of large models.
More recently, CoLLM~\cite{huynh2025collm} leverages LLMs both for triplet synthesis and for composed query embedding, introducing significant query-time latency.
ConText-CIR~\cite{xing2025context} employs cross-attention with a concept-consistency loss to align noun phrases in the modification text with relevant image regions.
Zero-shot CoIR methods~\cite{saito2023pic2word,tang2024context,lin2024fine,basic,jang2024slerp} have shown that manipulating representations in the embedding space can effectively perform CoIR without explicitly merging cross-modal data.
Pic2Word~\cite{saito2023pic2word}, Context-I2W~\cite{tang2024context}, and FTI4CIR~\cite{lin2024fine} introduced textual inversion to map an image into text tokens for later text-to-image retrieval.
BASIC~\cite{basic} introduces score-level late fusion of image-to-image and text-to-image similarities in a training-free manner, without relying on MLLMs at inference time.
However, BASIC shows low throughput due to the numerous text encoder forward passes.
Meanwhile, Slerp~\cite{jang2024slerp} employed spherical linear interpolation between visual and textual embeddings, showing that simple interpolation can yield competitive performance.
However, it employs a single dataset-level interpolation weight that overlooks sample-level diversity.
We address this issue by adaptively predicting a sample-wise, rank-aware interpolation weight.

\subsection{Composed Video Retrieval}
Composed Video Retrieval (CoVR)~\cite{webvid2024} aims to retrieve a target video given a reference video and a textual description of the intended modification.
\cite{webvid2024} introduced WebVid-CoVR, a large-scale dataset with video-text-video triplets derived from web video captions, and CoVR, a BLIP-based~\cite{li2022blip} model trained with contrastive learning~\cite{radenovic2023hnnce}.
CoVR-2~\cite{covr22024} enhanced CoVR by leveraging BLIP-2~\cite{li2023blip2} as a more capable multimodal backbone and introducing a caption-retrieval objective to provide auxiliary supervision.
\cite{thawakar2024composed} and \cite{simple2025} proposed frameworks that augment WebVid-CoVR with additional captions describing the difference between reference and target videos generated by an LLM, and use cross-attention mechanism to integrate video and textual representations within a unified grounding encoder.
TFR-CVR~\cite{hummel2024egocvr} adopted a training-free pipeline that combines LLM-generated captions with visual candidate filtering and text-based re-ranking, focusing on zero-shot generalization rather than end-to-end fusion learning.

While these approaches have advanced compositional understanding, they depend on large language models to fuse the reference and textual cues, which substantially increases query-time latency.
To overcome this limitation, our method adopts a lightweight and interpolation-based fusion that avoids introducing a heavy inference step or cross-modal modules.

\section{Proposed Method}
\label{sec:method}
This section presents details of our framework, SRAIN; its overall architecture is depicted in~\cref{fig:overall}.
We first describe the embedding extraction process for the reference and target visual data and the modification text in~\cref{sec:embedding_extraction}, and then elaborate on the rank-aware interpolation weight estimation strategy in~\cref{sec:optimal_interpolation}.
The two-stage training strategy of SRAIN is then introduced in~\cref{sec:two-stage}.
Finally,~\cref{sec:memory} describes the weight prediction procedure at inference time.

\subsection{Embedding Extraction}
\label{sec:embedding_extraction}
Following CoVR-2~\cite{covr22024}, SRAIN
employs BLIP-2~\cite{li2023blip2} as the pretrained vision-language model, which consists
of a ViT image encoder~\cite{ViT} and a Q-Former~\cite{li2023blip2} that projects visual and textual representations into a shared embedding space. %
In contrast to CoVR-2 that jointly processes the reference visual input and the modification text within BLIP-2,
SRAIN encodes each modality independently.
Extracting separate embeddings for the reference and modification allows efficient linear interpolation in the shared embedding space without introducing additional fusion modules.
More architectural details of our embedding networks are provided in the supplement.

\noindent\textbf{Reference Visual Embedding.}
A reference image is used as-is, and for a reference video, we use its middle frame as input following prior work~\cite{covr22024}.
The image is encoded by the frozen ViT encoder, producing patch embeddings that are subsequently queried by the Q-Former using $N_q$ learnable tokens.
The resulting
token outputs are passed through the BLIP-2 projection layer, averaged across tokens, and $\ell_2$-normalized to obtain the reference visual embedding $\mathbf{r}\in\mathbb{R}^{d}$.

\noindent\textbf{Modification Text Embedding.}
The modification text is tokenized and processed by the Q-Former without visual conditioning. The hidden state of the first token is projected by the BLIP-2 text projection layer and $\ell_2$-normalized to yield the modification text embedding $\mathbf{t} \in \mathbb{R}^d$.

\noindent\textbf{Target Visual Embedding.}
For a target image, we directly encode the image using the frozen ViT and Q-Former, producing $N_q$ token embeddings. These tokens are averaged and $\ell_2$-normalized to obtain the target visual embedding $\mathbf{v}\in\mathbb{R}^d$.
For a target video, we uniformly sample $N$ frames and encode each frame using the frozen ViT encoder.
The resulting patch embeddings are queried by the frozen Q-Former with $N_q$ learnable tokens, producing $N_q$ token embeddings per frame ($N \times N_q$ tokens in total).
To aggregate temporal information, query scoring~\cite{bain2022cliphitchhiker, covr22024} is applied separately for each query token: for each token, per-frame weights are computed based on the similarity between its embeddings across frames and the modification text embedding, and a weighted average over the frames produces a single temporally-aggregated token.
This yields $N_q$ aggregated tokens, which are then averaged and $\ell_2$-normalized to obtain the target visual embedding $\mathbf{v}\in\mathbb{R}^d$.
Since the BLIP-2 encoders remain frozen, these weights can be precomputed and cached.\footnote{Following CoVR-2~\cite{covr22024}, we freeze BLIP-2 for target visual data to reduce the computational cost for both training and inference.}

\subsection{Interpolation Weight Estimation}
\label{sec:optimal_interpolation}
SRAIN learns to predict sample-specific linear interpolation weights between reference visual embeddings and modification text embeddings to form composed query embeddings.
The major challenge in this direction is the absence of ground-truth interpolation weights.
The optimal interpolation weight must prioritize the discriminative power of the interpolated query embedding against negative samples, instead of merely maximizing its similarity to the true target embeddings.
In other words, the optimal weight is the one that achieves the most desirable ranking behavior, where positive targets are closer to the composed query than negative ones.
Hence, the search for the optimal interpolation weight involves comparison with all training data (or all data within the search database during testing), which results in a complexity proportional to the square of the dataset size, and thus determining the optimal interpolation weights is impractical.

To address this, we approximate the optimal weights within each mini-batch
of a sufficiently large size.
By leveraging hard negatives present in the batch,
we can effectively estimate near-optimal interpolation weights that reflect the rank-aware objective in practice.
The weight estimation procedure is illustrated in~\cref{fig:ranking}.
Given a batch of $B$ triplets, we consider a discrete set of $K$ interpolation candidates $\{\alpha^k\}_{k=1}^{K}$ uniformly spaced in $[0,1]$.
Inspired by previous work~\cite{jang2024slerp,shoemake1985animating}, we adopt spherical linear interpolation to fuse the reference visual and modification text embeddings.
For each triplet $i$ and candidate weight $\alpha^k$, the composed query embedding $\mathbf{f}_i^k$ is computed by
\begin{equation}
    \label{eq:lin}
    \mathbf{f}_i^k = \frac{\sin((1-\alpha^k)\theta_i)}{\sin(\theta_i)}\mathbf{r}_i + \frac{\sin(\alpha^k\theta_i)}{\sin(\theta_i)}\mathbf{t}_i,
\end{equation}
where $\theta_i$ is the angle between $\mathbf{r}_i$ and $\mathbf{t}_i$.
The fused query $\mathbf{f}_i^k$ is compared with all target embeddings $\{\mathbf{v}_j\}_{j\in{B}}$ in the batch using cosine similarity as follows:
\begin{equation}
    s_{i,j,k} =  \frac{ \mathbf{f}_i^{k\top} \mathbf{v}_j } {\|\mathbf{f}_i^k\|_2 \, \|\mathbf{v}_j\|_2}.
\end{equation}
For the $i$-th query, we construct a score matrix $\mathbf{S}_i=\big[s_{i,j,k}\big]_{B\times K}$ where each column corresponds to one interpolation candidate.

For each column $k$ of $\mathbf{S}_i$,
which corresponds to the similarity scores between the query fused with $\alpha^k$ and all targets in the batch,
$\mathrm{rank}_i^k$ denote the rank of the positive target ($j=i$) among the $B$ similarity scores after sorting them in descending order.
In other words, $\mathrm{rank}_i^k \in \{1, \dots, B\}$ denotes the index of the positive target in the sorted list, where a smaller value indicates better retrieval performance.
The approximate optimal interpolation weight for the $i$-th query is then selected as
\begin{equation}
    \label{eq:alpha}
    \alpha_i^{*} :=\alpha^z, \textrm{where}\; z=\underset{k \in [1,K]}{\mathrm{argmin}}\;\mathrm{rank}_i^k.
\end{equation}
In cases where multiple candidates achieve the minimum rank, we set $\alpha_i^{*}$ to their median.
The selected $\alpha_i^{*}$ then serves as the ground-truth label.

\subsection{Two-stage Training Strategy}
\label{sec:two-stage}
Training
SRAIN is conducted in two stages.
In the first stage, the Q-Former~\cite{li2023blip2} is fine-tuned with the ground-truth weights to achieve \emph{modality alignment}, \ie, aligning the fused query embeddings with their corresponding target embeddings while pushing apart negatives. %
In the second stage, the learned Q-Former is frozen, and a lightweight
module for interpolation weight prediction is trained to regress the
ground-truth weights.
This two-stage design is motivated by the need to freeze the encoder during the weight predictor training.
Since the ground-truth weights are computed based on ranking behavior within the feature space of the encoder, updating the encoder simultaneously with the weight predictor causes the estimated weights to vary even for the same query pair, making it impossible to provide consistent supervision to the weight predictor.

Following~\cref{eq:lin}, the fused query embedding $\mathbf{f}_i\in\mathbb{R}^d$ for each triplet $i$ is obtained by spherical linear interpolation between the reference visual embedding $\mathbf{r}_i$ and the modification text embedding $\mathbf{t}_i$ using the ground-truth interpolation weight $\alpha_i^*$.
The goal of the first stage is to align $\mathbf{f}_i$ with its corresponding target visual embedding $\mathbf{v}_i\in\mathbb{R}^d$. To this end, we employ the hard-negative contrastive loss~\cite{radenovic2023hnnce}, which encourages higher similarity for positive pairs while suppressing hard negatives through adaptive weighting based on their hardness.
Given a batch with $B$ triplets, the loss is formulated as:
\begin{equation}
    \begin{aligned}
        \mathcal{L}(\mathcal{B}) = & -\sum_{i=1}^{B}\Bigg(\log\frac{e^{s_{ii}/\tau }}{\gamma \cdot e^{s_{ii}/\tau}+\sum_{j=1, j\neq i}^{B}e^{s_{ij}/\tau }w_{ij}} \Bigg) \\
                                   & -\sum_{i=1}^{B}\Bigg(\log\frac{e^{s_{ii}/\tau }}{\gamma \cdot e^{s_{ii}/\tau}+\sum_{j=1, j\neq i}^{B}e^{s_{ji}/\tau}w_{ji}} \Bigg),
        \label{eq:loss_align}
    \end{aligned}
\end{equation}
where $s_{ij}$ denotes the cosine similarity between the fused query embedding $\mathbf{f}_i$ and the target visual embedding $\mathbf{v}_j$.

In the second stage, the interpolation weight predictor is trained to predict the ground-truth interpolation weight $\alpha_i^*$ given the query inputs while the pretrained encoders are fixed.
Predicting the interpolation weight solely from the reference visual embedding and the modification text embedding is challenging since the ground-truth weight is estimated in a rank-aware manner that explicitly depends on hard negatives within the batch.
To address this, as illustrated in~\cref{fig:overall},  we introduce a conditioned weight prediction mechanism that enables the weight predictor to infer more accurate weights by conditioning on informative negative samples.
Specifically,
the weight predictor takes as auxiliary input a conditioning token set $C_i$ that consists of the positive target embedding $\mathbf{v}_i$ and $m-1$ hardest negative target embeddings, \ie, those of negative targets most similar with the fused query embedding $\mathbf{f}_i$.

The interpolation weight predictor consists of a transformer encoder~\cite{vaswani2017attention} with $L$ layers followed by a three-layer MLP.
It takes as input the reference embedding $\mathbf{r}_i^\top\in\mathbb{R}^{1\times d}$, the modification embedding $\mathbf{t}_i^\top\in\mathbb{R}^{1\times d}$, the conditioning token set $C_i\in\mathbb{R}^{m\times d}$, and a learnable prediction token $\mathbf{p}\in\mathbb{R}^{1\times d}$.
The prediction token aggregates information from all inputs through the transformer encoder to produce an output embedding:
\begin{equation}
    \label{eq:predictor}
    \mathbf{p}' = \texttt{Transformer}\big([\mathbf{r}_i^\top,\, \mathbf{t}_i^\top,\, C_i,\, \mathbf{p}];\theta_{t}\big),
\end{equation}
where $\theta_t$ denotes the encoder parameters and $\mathbf{p}' \in \mathbb{R}^{1\times d}$ is the output embedding.
This embedding is then fed into the MLP, which projects it to a scalar and applies a sigmoid function to produce the predicted weight:
\begin{equation}
    \hat{\alpha}_i = \sigma\big(\texttt{MLP}(\mathbf{p}')\big).
\end{equation}
The details of the architecture of the predictor are presented in the supplementary material.
The predictor is optimized by minimizing the mean squared error between the predicted and the ground-truth interpolation weights as
\begin{equation}
    \mathcal{L}_{\text{reg}} = \frac{1}{B}\sum_{i=1}^{B}(\hat{\alpha}_i - \alpha_i^*)^2.
\end{equation}

\subsection{Weight Prediction at Inference} %
\label{sec:memory}
At inference time, the positive and negative target embeddings are unavailable, preventing direct construction of the conditioning token set.
To address this, we introduce a memory bank $\mathcal{M} = \{\mathbf{m}_k\}_{k=1}^{|\mathcal{M}|}$ of prototype embeddings $\mathbf{m}_k \in\mathbb{R}^{d}$ that represent the distribution of target visual data for training. During training, each prototype is updated by exponential moving average with momentum $\lambda$:
\begin{equation}
    \label{eq:ema}
    \mathbf{m}_z \leftarrow \lambda \cdot \mathbf{m}_z + (1-\lambda) \cdot \mathbf{v}_i,
    \;
    z = \underset{k\in[1,|\mathcal{M}|]}{\mathrm{argmax}} \frac{\mathbf{m}_k^{\top}\mathbf{v}_i}{\|\mathbf{m}_k\|_2 \|\mathbf{v}_i\|_2}.
\end{equation}
After training, the memory bank is frozen and used during inference to provide conditioning tokens for the weight predictor.
To select informative prototypes regardless of the fusion weight, we fuse the query with multiple candidate interpolation weights uniformly sampled from $[0,1]$, similar to~\cref{sec:optimal_interpolation}.
For each candidate weight, we retrieve the top-$k$ most similar prototypes from the memory bank.
We then select the $n$ prototypes that are most frequently retrieved across all candidate weights, which are likely to be hard negatives.
This approach maintains consistency between training and inference by providing informative reference points that enable accurate weight prediction without access to batch negatives.
The justification for this approximation and detailed analysis of the memory bank's capacity are provided in the supplement.

\section{Experiments}
\label{sec:experiments}
This section first describes the experimental setup, including datasets, evaluation metrics, and implementation details in~\cref{sec:ex_setup}.
Then \cref{sec:quant} provides quantitative results to demonstrate the effectiveness of SRAIN compared to previous work.
Next, we include the qualitative results of SRAIN in~\cref{sec:qual}, and finally analyze the impact of SRAIN's components through ablation studies in~\cref{sec:ablation}.

\subsection{Experimental Setup}
\label{sec:ex_setup}
\noindent\textbf{Datasets.}
We evaluate and compare the performance of our method against previous work on one CoVR benchmark, WebVid-CoVR~\cite{webvid2024}, and three CoIR benchmarks, FashionIQ~\cite{FIQ}, CIRCO~\cite{CIRCO} and CIRR~\cite{CIRR}.
\textbf{WebVid-CoVR} is a large-scale CoVR dataset containing 1.64M video-text-video triplets for training and 2,556 triplets for testing, derived from web video captions. Each triplet consists of a reference video, a textual modification describing the desired change, and a target video.
\textbf{FashionIQ} is a CoIR dataset in the fashion domain, containing 30,134 triplets from 77,684 images across three categories: Dress, Shirt, and Toptee.
The images are split into 6:2:2 for training, validation, and test.
Each query pairs a reference image with natural language feedback describing desired modifications such as changes in style, color, or pattern.
We report the validation recalls because the labels of the evaluation split have not been publicly released.
\textbf{CIRCO} is a zero-shot CoIR benchmark built on COCO images, containing 800 test queries with a retrieval gallery size of 123,403 for evaluation.
The benchmark is specifically designed to assess model generalization capability with test queries.
We follow the standard zero-shot evaluation protocol, where models are not fine-tuned on CIRCO.
\textbf{CIRR} is a CoIR dataset sharing the same triplet structure as FashionIQ, but covering open-domain scenes rather than a specific category. It is built on real-life images sourced from the NLVR2 dataset, containing 21,552 images and 36,554 triplets split in an 8:1:1 ratio for training, validation, and test.

\begin{table*}[!t]
    \fontsize{7}{8.5}\selectfont
    \centering
    \caption{
        Composed video retrieval results on the WebVid-CoVR-Test, where all compared methods are fine-tuned on the WebVid-CoVR training set.
        Compared methods differ in fusion strategy: Avg indicates element-wise averaging of reference and modification embeddings, MLP denotes a multi-layer perceptron-based fusion module, CA employs cross-attention, and LI refers to
        linear interpolation.
        External Knowledge denotes the use of external models
        for additional video descriptions generation.
    }
    \vspace{-2ex}
    \begin{tabularx}{1.0\textwidth}
        {
        >{\raggedright\arraybackslash}p{0.22\textwidth}
        >{\centering\arraybackslash}p{0.143\textwidth}
        >{\centering\arraybackslash}p{0.147\textwidth}
        >{\centering\arraybackslash}p{0.072\textwidth}
        >{\centering\arraybackslash}p{0.115\textwidth}
        >{\centering\arraybackslash}X
        >{\centering\arraybackslash}X
        >{\centering\arraybackslash}X
        >{\centering\arraybackslash}X
        }
        \toprule
        {\multirow{1}{*}[-4.4mm]{\textbf{Methods}}} & \multirow{1}{*}[-1.5mm]{\textbf{External}}     & \multirow{1}{*}[-4.4mm]{\textbf{Modality}} & \multirow{1}{*}[-4.4mm]{\textbf{Fusion}} & \multirow{1}{*}[-4.4mm]{\textbf{Backbone}}
                                                    & \multicolumn{4}{c}{\textbf{Evaluation Metric}}                                                                                                                                                                                                                       \\ [-0.3ex]  \cmidrule(lr){6-9}
                                                    & \multirow{1}{*}[-0.0mm]{\textbf{Knowledge}}    &                                            &                                          &                                            & R@1                  & R@5                   & R@10           & R@50           \\  [-0.0ex] \midrule
        Text-only ($\alpha=1$)                      & \xmark                                         & Text                                       & -                                        & BLIP-2                                     & 23.32                & 46.21                 & 56.22          & 78.36          \\
        Visual-only ($\alpha=0$)                    & \xmark                                         & Visual                                     & -                                        & BLIP-2                                     & 36.03                & 64.24                 & 74.77          & 92.64          \\
        Average ($\alpha=0.5$)                      & \xmark                                         & Visual+Text                                & Avg                                      & BLIP-2                                     & 58.69                & 83.88                 & 90.49          & 98.21          \\
        CoVR~\cite{webvid2024}                      & \xmark                                         & Visual+Text                                & MLP                                      & BLIP                                       & 50.59                & 74.65                 & 83.57          & 95.46          \\
        CoVR-2~\cite{covr22024}                     & \xmark                                         & Visual+Text                                & MLP                                      & BLIP-2                                     & 51.88                & 79.38                 & 86.46          & 97.42          \\
        CoVR~\cite{webvid2024}                      & \xmark                                         & Visual+Text                                & CA                                       & BLIP                                       & 55.95                & 81.22                 & 89.05          & 98.08          \\
        CoVR-2~\cite{covr22024}                     & \xmark                                         & Visual+Text                                & CA                                       & BLIP-2                                     & 59.82                & 83.84                 & \textbf{91.28} & 98.24          \\
        Thawakar~\etal~\cite{thawakar2024composed}  & MiniGPT4~\cite{zhu2023minigpt}                 & Visual+Text                                & CA                                       & BLIP                                       & 60.12                & 84.32                 & 91.27          & \textbf{98.72} \\
        \ccol  SRAIN (Ours)                         & \ccol \xmark                                   & \ccol  Visual+Text                         & \ccol LI                                 & \ccol  BLIP-2                              & \ccol \textbf{61.07} & \ccol  \textbf{84.90} & \ccol 91.20    & \ccol 98.40    \\
        \bottomrule
    \end{tabularx}

    \label{tab:main_results}
    \vspace{-2ex}
\end{table*}

\noindent\textbf{Evaluation Metrics.}
We employ the standard metric of recall at $K$ (R@$K$) to evaluate retrieval performance for both composed video retrieval and composed image retrieval.
Specifically, we report R@$K$ with $K$=1, 5, 10, 50 for WebVid-CoVR, $K$=10, 50 for FashionIQ and $K$=1, 5, 10 for CIRR.
In all datasets, samples are ranked based on their similarities to the query.
For CIRCO, we follow the protocol of Baldrati~\etal~\cite{CIRCO} and evaluate our method with a ranking-based metric, mean average precision at $K$ (mAP@$K$, with $K$=5, 10, 25, 50).

\noindent\textbf{Implementation Details.}
We use pretrained BLIP-2~\cite{li2023blip2} with ViT-G/14~\cite{ViT} and Q-Former~\cite{li2023blip2} as our vision-language backbone following CoVR-2~\cite{covr22024}.\linebreak
SRAIN is trained for 5 epochs for each stage with batch size 512 using AdamW optimizer~\cite{adamw} with learning rates of $\expnum{2}{5}$ for Q-Former and $\expnum{1}{3}$ for other parameters.
The interpolation weight predictor comprises a 2-layer Transformer encoder and a 3-layer MLP, and contains 1.63M parameters, which corresponds to only 0.14\% of the total model size.
During training, it takes 50 conditioning tokens consisting of 1 positive and 49 batch-level hard negatives.
Since the hard negatives are only processed by this lightweight predictor, they introduce a negligible computational overhead of merely 0.03 ms per query, accounting for 0.08\% of the total wall-clock time.
We set $|\mathcal{A}|=101$ in~\cref{sec:optimal_interpolation}.
The memory bank has 1,024 prototypes and is updated by $\lambda=0.99$ in~\cref{eq:ema}.
For the hard-negative contrastive loss in~\cref{eq:loss_align}, we set $\gamma=1$, $\tau=0.07$, and compute $w_{ij}$ following~\cite{radenovic2023hnnce} with $\beta=0.5$.
These settings apply to WebVid-CoVR~\cite{webvid2024}, FashionIQ~\cite{FIQ} and CIRR~\cite{CIRR}.
More implementation details are presented in the supplement.

\begin{table*}[!t]
    \fontsize{6.}{7.5}\selectfont
    \centering
    \caption{Composed image retrieval results on the FashionIQ validation set.
        CC3M~\cite{sharma2018conceptual}, LAION2B (L2B)~\cite{schuhmann2022laion}, LAION2M (L2M), and LLaVA-Align (L-A)~\cite{llava} contain 3M, 2.3B, 2M, and 585k image-text pairs, respectively, where LAION2M is a subset of LAION400M~\cite{schuhmann2021laion}.
        MagicLens36M~\cite{zhang2024magiclens} comprises 36M (query image, instruction, target image) triplets.
        SDP~\cite{zhang2023_stable_diffusion_prompts_247m} provides 2.47M text prompts, while ST18M~\cite{gu2024compodiff} and LaSCo~\cite{levy2024data} comprise 18.8M and 389k triplets for training, respectively.
    }
    \vspace{-1ex}
    \begin{tabularx}{1.0\textwidth}
        {
            >{\raggedright\arraybackslash}p{0.16\textwidth}
            >{\centering\arraybackslash}X
            >{\centering\arraybackslash}p{0.12\textwidth}
            >{\centering\arraybackslash}p{0.17\textwidth}
            >{\centering\arraybackslash}X
            >{\centering\arraybackslash}X
            >{\centering\arraybackslash}X
            >{\centering\arraybackslash}X
            >{\centering\arraybackslash}X
            >{\centering\arraybackslash}X
            >{\centering\arraybackslash}X
            >{\centering\arraybackslash}X
        }
        \toprule
        {\multirow{1}{*}[-4.mm]{\textbf{Methods}}} & {\multirow{1}{*}[-4.mm]{\textbf{ViT}}} & \multirow{1}{*}[-1.5mm]{\textbf{External}} & \multirow{1}{*}[-1.5mm]{\textbf{Pretraining}} & \multicolumn{2}{c}{\textbf{Dress}} & \multicolumn{2}{c}{\textbf{Shirt}} & \multicolumn{2}{c}{\textbf{Toptee}} & \multicolumn{2}{c}{\textbf{Average}}                                                                                              \\ [-0.3ex]  \cmidrule(lr){5-6} \cmidrule(lr){7-8} \cmidrule(lr){9-10} \cmidrule(lr){11-12}

                                                   &                                        & {\textbf{Knowledge}}                       & \textbf{Data}                                 & R@10                               & R@50                               & R@10                                & R@50                                 & R@10                 & R@50                 & R@10                 & R@50                  \\  [-0.4ex] \midrule

        \multicolumn{12}{l}{\textit{\textbf{Zero-shot}}}                                                                                                                                                                                                                                                                                                                                                                                     \\ [-0.3ex]\midrule
        CIReVL~\cite{cirevl}                       & G                                      & GPT-4~\cite{achiam2023gpt4}                & \xmark                                        & 27.07                              & 49.53                              & 33.71                               & 51.42                                & 35.80                & 56.14                & 32.19                & 52.36                 \\
        OSrCIR~\cite{tang2025reason}               & G                                      & GPT-4o~\cite{achiam2023gpt4}               & \xmark                                        & 33.02                              & 54.78                              & 38.65                               & 54.71                                & 41.04                & 61.83                & 37.57                & 57.11                 \\ \cmidrule(lr){1-12}
        Pic2Word~\cite{saito2023pic2word}          & L                                      & \xmark                                     & CC3M                                          & 20.00                              & 40.20                              & 26.20                               & 43.60                                & 27.90                & 47.40                & 24.70                & 43.70                 \\
        SEARLE-XL~\cite{CIRCO}                     & L                                      & \xmark                                     & CC3M                                          & 26.89                              & 45.58                              & 20.48                               & 43.13                                & 29.32                & 49.97                & 25.56                & 46.23                 \\
        TAT~\cite{jang2024slerp}                   & L                                      & \xmark                                     & CC3M+L2M+L-A                                  & 29.15                              & 50.62                              & 32.14                               & 51.62                                & 37.02                & 57.73                & 32.77                & 53.32                 \\
        MagicLens~\cite{zhang2024magiclens}        & L                                      & \xmark                                     & MagicLens36M                                  & 25.50                              & 46.10                              & 32.70                               & 53.80                                & 34.00                & 57.70                & 30.70                & 52.50                 \\
        LinCIR~\cite{gu2024language}               & G                                      & \xmark                                     & CC3M+SDP                                      & 38.08                              & 60.88                              & 46.76                               & 65.11                                & 50.48                & 71.09                & 45.11                & 65.69                 \\ [-0.4ex]  \cmidrule(lr){1-12}
        CoVR~\cite{webvid2024}                     & L                                      & \xmark                                     & WebVid-CoVR                                   & 21.95                              & 39.05                              & 30.37                               & 46.12                                & 30.78                & 48.73                & 27.70                & 44.63                 \\
        CoVR-2~\cite{covr22024}                    & G                                      & \xmark                                     & WebVid-CoVR                                   & -                                  & -                                  & -                                   & -                                    & -                    & -                    & 27.78                & -                     \\
        \ccol  SRAIN (Ours)                        & \ccol G                                & \ccol \xmark                               & \ccol WebVid-CoVR                             & \ccol \textbf{23.22}               & \ccol  \textbf{43.33}              & \ccol \textbf{33.14}                & \ccol  \textbf{52.01}                & \ccol \textbf{31.91} & \ccol \textbf{49.03} & \ccol \textbf{29.42} & \ccol  \textbf{48.12} \\ \midrule
        \multicolumn{12}{l}{\textit{\textbf{Fine-tuning on FashionIQ}}}                                                                                                                                                                                                                                                                                                                                                                      \\ [-0.3ex]\midrule
        CompoDiff~\cite{gu2024compodiff}           & G                                      & \xmark                                     & ST18M+L2B                                     & 38.39                              & 51.03                              & 41.68                               & 56.02                                & 45.70                & 57.32                & 39.81                & 51.90                 \\
        CoVR~\cite{webvid2024}                     & L                                      & \xmark                                     & WebVid-CoVR                                   & 44.55                              & 69.03                              & 48.43                               & 67.42                                & 52.60                & 74.31                & 48.53                & 70.25                 \\
        CASE~\cite{levy2024data}                   & G                                      & \xmark                                     & LaSCo                                         & 47.44                              & 69.36                              & 48.48                               & 70.23                                & 50.18                & 72.24                & 48.79                & 70.68                 \\ [-0.4ex]  \cmidrule(lr){1-12}%
        MAAF~\cite{dodds2020modality}              & L                                      & \xmark                                     & \xmark                                        & 23.80                              & 48.60                              & 21.30                               & 44.20                                & 27.90                & 53.60                & 24.30                & 48.80                 \\
        LF-BLIP~\cite{levy2024data}                & L                                      & \xmark                                     & \xmark                                        & 25.31                              & 44.05                              & 25.39                               & 43.57                                & 26.54                & 44.48                & 25.75                & 43.98                 \\
        CoSMo~\cite{lee2021cosmo}                  & L                                      & \xmark                                     & \xmark                                        & 25.64                              & 50.30                              & 24.90                               & 49.18                                & 29.21                & 57.46                & 26.58                & 52.31                 \\
        DCNet~\cite{kim2021dual}                   & L                                      & \xmark                                     & \xmark                                        & 28.95                              & 56.07                              & 23.95                               & 47.30                                & 30.44                & 58.29                & 27.78                & 53.89                 \\
        FashionVLP~\cite{goenka2022fashionvlp}     & L                                      & \xmark                                     & \xmark                                        & 32.42                              & 60.29                              & 31.89                               & 58.44                                & 38.51                & 68.79                & 34.27                & 62.51                 \\
        CLIP4CIR~\cite{clip4cir}                   & L                                      & \xmark                                     & \xmark                                        & 31.63                              & 56.67                              & 36.36                               & 58.00                                & 38.19                & 62.42                & 35.39                & 59.03                 \\
        CoVR-2~\cite{covr22024}                    & G                                      & \xmark                                     & \xmark                                        & 45.25                              & 68.86                              & 49.95                               & 69.95                                & \textbf{51.37}       & \textbf{72.56}       & 48.86                & \textbf{70.46}        \\
        \ccol  SRAIN (Ours)                        & \ccol G                                & \ccol \xmark                               & \ccol \xmark                                  & \ccol \textbf{46.25}               & \ccol  \textbf{69.07}              & \ccol \textbf{51.23}                & \ccol  \textbf{70.00}                & \ccol 50.30          & \ccol 71.75          & \ccol \textbf{49.26} & \ccol  70.27          \\
        \bottomrule
    \end{tabularx}

    \label{tab:fiq}
    \vspace{-3ex}
\end{table*}

\subsection{Quantitative Results}
\label{sec:quant}
\begin{table}[!t]
    \fontsize{8}{9}\selectfont
    \centering
    \caption{Zero-shot composed image retrieval results on CIRCO.}
    \vspace{-1ex}
    \begin{tabularx}{0.75\textwidth}
        {
        >{\raggedright\arraybackslash}p{0.18\textwidth}
        >{\centering\arraybackslash}X
        >{\centering\arraybackslash}p{0.15\textwidth}
        >{\centering\arraybackslash}X
        >{\centering\arraybackslash}X
        >{\centering\arraybackslash}X
        >{\centering\arraybackslash}X
        }
        \toprule
        {\multirow{1}{*}[-4.5mm]{\textbf{Methods}}} & {\multirow{1}{*}[-4.5mm]{\textbf{ViT}}} & \multirow{1}{*}[-1.5mm]{\textbf{External}} & \multicolumn{4}{c}{\textbf{mAP@$K$}}                                                                      \\ [-0.3ex]  \cmidrule(lr){4-7}
                                                    &                                         & \textbf{Knowledge}                         & $K$=5                                & $K$=10               & $K$=25               & $K$=50               \\  [-0.4ex] \midrule
        TFCIR~\cite{sun2023training}                & G                                       & \cmark                                     & 26.52                                & 28.25                & 31.23                & 31.99                \\
        CIReVL~\cite{cirevl}                        & G                                       & \cmark                                     & 26.77                                & 27.59                & 29.96                & 31.03                \\
        OSrCIR~\cite{tang2025reason}                & G                                       & \cmark                                     & 30.47                                & 31.14                & 35.03                & 36.59                \\ \cmidrule(lr){1-7}
        SEARLE-XL~\cite{CIRCO}                      & L                                       & \xmark                                     & 11.68                                & 12.73                & 14.33                & 15.12                \\
        CompoDiff~\cite{gu2024compodiff}            & G                                       & \xmark                                     & 15.33                                & 17.71                & 19.45                & 21.01                \\
        LinCIR~\cite{gu2024language}                & G                                       & \xmark                                     & 19.71                                & 21.01                & 23.13                & 24.18                \\
        BASIC~\cite{basic}                          & G                                       & \xmark                                     & 19.98                                & 20.88                & 22.85                & 23.71                \\
        CoVR~\cite{webvid2024}                      & L                                       & \xmark                                     & 21.43                                & 22.33                & 24.47                & 25.48                \\
        CoVR-2~\cite{covr22024}                     & G                                       & \xmark                                     & 25.06                                & -                    & -                    & -                    \\

        \ccol  SRAIN (Ours)                         & \ccol  G                                & \ccol  \xmark                              & \ccol \textbf{26.16}                 & \ccol \textbf{27.34} & \ccol \textbf{29.94} & \ccol \textbf{30.81} \\
        \bottomrule
    \end{tabularx}

    \label{tab:circo}
    \vspace{-4ex}
\end{table}

\noindent\textbf{Composed Video Retrieval.}
We evaluate SRAIN on the WebVid-CoVR-Test benchmark and compare it with state-of-the-art methods in~\cref{tab:main_results}.
SRAIN surpasses all existing methods in terms of R@1.
Notably, SRAIN outperforms~\cite{thawakar2024composed}, which uses cross-attention fusion and external MLLM-generated captions, by 0.95\% in R@1.
Compared to CoVR~\cite{webvid2024} and CoVR-2~\cite{covr22024} that adopt cross-attention fusion, SRAIN achieves significant improvements of 5.12\% and 1.25\% in R@1, respectively, using a simple but effective linear interpolation strategy.
These results demonstrate that SRAIN enables effective multimodal fusion without relying on computationally expensive cross-attention or external models.

\begin{wraptable}{r}{0.49\textwidth}
    \vspace{-5ex}
    \fontsize{8}{9.5}\selectfont
    \centering
    \caption{Composed image retrieval results on CIRR.}
    \begin{tabularx}{0.49\textwidth}
        {
        >{\raggedright\arraybackslash}p{0.18\textwidth}
        >{\centering\arraybackslash}X
        >{\centering\arraybackslash}X
        >{\centering\arraybackslash}X
        >{\centering\arraybackslash}X
        }
        \toprule
        {\multirow{1}{*}[-4.5mm]{\textbf{Methods}}} & {\multirow{1}{*}[-4.5mm]{\textbf{ViT}}} & \multicolumn{3}{c}{\textbf{R@$K$}}                                               \\ [-0.3ex]  \cmidrule(lr){3-5}
                                                    &                                         & $K$=1                              & $K$=5                & $K$=10               \\  [-0.4ex] \midrule
        CLIP4CIR~\cite{clip4cir}                    & L                                       & 33.59                              & 65.35                & 77.35                \\
        CompoDiff~\cite{gu2024compodiff}            & G                                       & 32.39                              & 57.61                & 77.25                \\
        CoVR~\cite{webvid2024}                      & L                                       & 49.69                              & 78.60                & 86.77                \\
        CoVR-2~\cite{covr22024}                     & G                                       & 50.87                              & 80.80                & 88.84                \\
        \ccol  SRAIN (Ours)                         & \ccol  G                                & \ccol \textbf{50.95}               & \ccol \textbf{81.21} & \ccol \textbf{89.60} \\
        \bottomrule
    \end{tabularx}
    \label{tab:cirr}
    \vspace{-2ex}
\end{wraptable}

\noindent\textbf{Composed Image Retrieval.}
We further assess SRAIN on FashionIQ~\cite{FIQ}, CIRCO~\cite{CIRCO} and CIRR~\cite{CIRR}, as summarized in~\cref{tab:fiq},~\cref{tab:circo} and~\cref{tab:cirr}, respectively.
In all tables, we explicitly report the ViT backbone size for each method, where L and G denote ViT-L and ViT-G, respectively. For methods appearing in both tables, the backbone and pretraining data in~\cref{tab:circo} are identical to those reported in~\cref{tab:fiq}.
On FashionIQ, SRAIN is evaluated in both the zero-shot and fine-tuned settings.
Our model trained solely on WebVid-CoVR achieves competitive performance in the zero-shot setting.
While most existing methods show strong results, their zero-shot performance is largely attributed to pretraining on significantly larger image–text corpora or leveraging external knowledge.
In contrast, our zero-shot evaluation is based on a model trained with the same pretraining data used for composed video retrieval.
We note that WebVid-CoVR consists solely of web video triplets with limited diversity and does not inherently favor out-of-domain image benchmarks.
The strong zero-shot results reflect the genuine generalization capability of SRAIN rather than data-related confounders.
For fine-tuning, SRAIN outperforms previous methods in average R@10.
On CIRCO, SRAIN outperforms all methods that do not use external knowledge in all metrics, and remains competitive with approaches that rely on external knowledge and incur heavy computational cost.
On CIRR, where all methods are evaluated under the supervised setting, SRAIN surpasses all methods across all retrieval metrics.

\subsection{Qualitative Results}
\label{sec:qual}
\begin{figure*}[t!]
    \centering
    \includegraphics[width=0.99\linewidth]{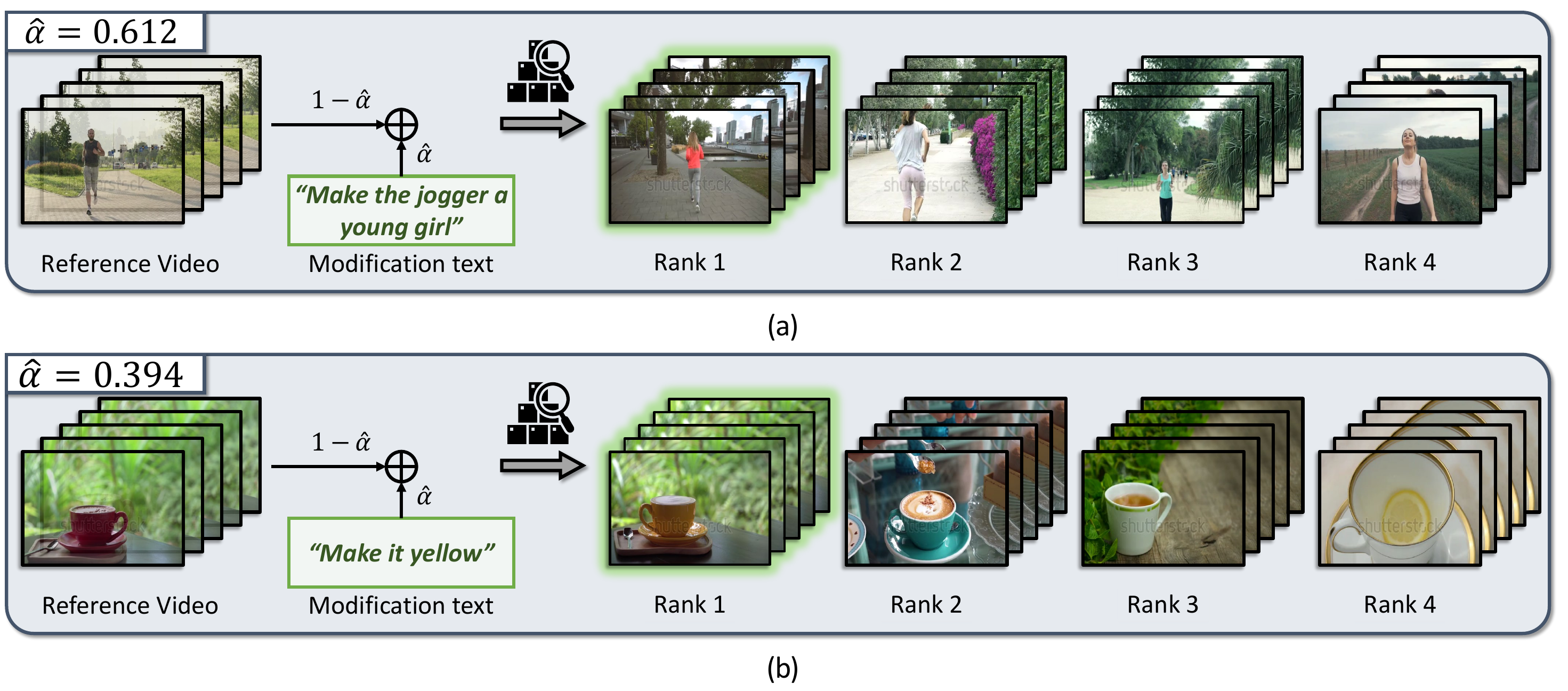}
    \caption{Top-4 retrieval results on the WebVid-CoVR test set.
        The ground-truth target video is highlighted in \color{grn}{green}.
    }
    \label{fig:qual}
    \vspace{-4ex}
\end{figure*}

\cref{fig:qual} illustrates the Top-4 retrieved video results from our method for the given query.
In~\cref{fig:qual}(a), the modification text ``\sftype{Make the jogger a young girl}'' is highly discriminative and sufficient to retrieve the correct target, where the predicted weight $\hat{\alpha}$ is relatively high, focusing more on the modification.
In contrast,~\cref{fig:qual}(b) presents a case where the text ``\sftype{Make it yellow}'' is visually ambiguous, leading the model to assign a lower $\hat{\alpha}$ and rely more on the reference video to retrieve the correct target.
These results highlight the benefit of our framework in adaptively controlling the influence of visual and textual cues, enabling precise and context-aware fusion tailored to each query.

\subsection{Ablation Studies}
\label{sec:ablation}
We evaluate the effectiveness of the proposed components in SRAIN through comprehensive experiments.
For the ablation study, we report composed video retrieval results on the WebVid-CoVR~\cite{webvid2024}.

\noindent\textbf{Effect of two-stage training strategy.}
We examine the effect of training strategy, as presented in~\cref{tab:ablation}~(a).
With the frozen BLIP-2, training the weight predictor alone (``Stage 2 only'') notably improves the performance over averaging (``Average''), demonstrating the effectiveness of our sample-wise interpolation approach even without encoder fine-tuning, while exhibiting lower performance overall compared to our full two-stage approach; it highlights the necessity of stage 1 for learning accurate cross-modal alignment.
On the other hand, end-to-end training underperforms the proposed two-stage strategy despite jointly optimizing the encoders and the predictor.
This result supports our design choice, as end-to-end training leads to unstable training since the ground-truth interpolation weight changes as the encoder updates, degrading the consistency of the supervision.
Meanwhile, we also include a ``Stage 1 only'' baseline, where the Q-former is fine-tuned with estimated optimal weights ($\alpha^*$) and a fixed weight ($\alpha=0.5$) is used for evaluation without the weight predictor, since $\alpha = 0.5$ is empirically identified as the globally optimal fixed interpolation weight.
Note that this comparison favors ``Stage 1 only'' because it uses the per-dataset globally optimal weight, whereas SRAIN predicts weights autonomously at test time.
Since the best global $\alpha$ is highly dataset-dependent~\cite{jang2024slerp} and performance is highly sensitive to the choice of $\alpha$, the gains from our weight predictor are substantial.
Moreover, the ``Stage 2 only'' variant provides a clear boost over the frozen baseline, underscoring SRAIN's value for zero-shot or data-constrained applications.
The full ablation across all benchmarks and both BLIP-2 and BLIP backbones is provided in the supplement, confirming that these improvements are consistent.

\begin{table}[!t]
    \fontsize{8}{9}\selectfont
    \centering
    \caption{Ablation studies on key components of our method.}
    \label{tab:ablation}
    \begin{tabularx}{0.70\textwidth}
        {
        >{\raggedright\arraybackslash}p{0.35\textwidth}
        >{\centering\arraybackslash}X %
        >{\centering\arraybackslash}X%
        >{\centering\arraybackslash}X%
        >{\centering\arraybackslash}X%
        }
        \toprule
        \textbf{Methods}                      & R@1         & R@5         & R@10        & R@50        \\  [-0.4ex] \midrule
        \multicolumn{5}{l}{\textbf{(a) \textit{Training Strategy}}}                                   \\ [-0.3ex]\midrule
        Average (BLIP-2 frozen)               & 45.66       & 71.71       & 81.30       & 94.80       \\
        Stage 2 only (BLIP-2 frozen)          & 48.08       & 73.40       & 83.22       & 95.74       \\
        End-to-end training                   & 57.36       & 82.71       & 89.20       & 97.77       \\
        Stage 1 only                          & 60.50       & 83.90       & 91.09       & 98.40       \\
        \midrule
        \multicolumn{5}{l}{\textbf{(b) \textit{Effect of $C_i$ in~\cref{eq:predictor}}}}              \\ [-0.3ex]\midrule
        w/o $C_i$ (Train \& Test)             & 60.33       & 84.66       & 90.92       & 98.36       \\
        w/o Memory $\mathcal{M}$ (Train only) & 60.62       & 84.65       & 91.09       & 98.40       \\ \midrule
        \multicolumn{5}{l}{\textbf{(c) \textit{Oracle (Upper bound)}}}                                \\ [-0.3ex]\midrule
        Stage 1 + Optimal $\alpha^*$          & 71.95       & 89.95       & 94.41       & 99.10       \\ \midrule
        \ccol  SRAIN (Ours)                   & \ccol 61.07 & \ccol 84.90 & \ccol 91.20 & \ccol 98.40 \\
        \bottomrule
    \end{tabularx}
    \vspace{-4ex}
\end{table}

\noindent\textbf{Effect of conditioned weight prediction mechanism.}
We further evaluate the impact of the conditioned prediction mechanism in~\cref{tab:ablation}~(b), showing that using $C_i$ during training improves performance over not using it at all.
Introducing the memory bank at test time to maintain train–test consistency further improves the performance.

\noindent\textbf{Upper bound analysis.}
After stage 1, we report an oracle performance by assigning the optimal interpolation weights computed via the rank-aware weight estimation procedure in~\cref{sec:optimal_interpolation} across the entire test set~(\cref{tab:ablation}(c)).
It achieves remarkable performance, suggesting that sample-aware weighting can drive significant performance gains, and that further improvement is possible through better prediction of instance-specific interpolation weights.
The gap to the upper bound is mainly attributed to the predictor's difficulty in approximating extreme values in the tails of the optimal weight distribution.
While the mean prediction error is near zero, indicating no systematic bias, the nontrivial standard deviation reveals that errors concentrate in these extreme cases.
A detailed analysis of the oracle weight distribution over the test set is included in the supplement.

\begin{table}[!t]
    \centering
    \caption{Ablation studies on the memory bank for zero-shot CoIR.}
    \label{tab:memory_bank}
    \vspace{-1ex}
    \fontsize{8}{9}\selectfont
    \begin{tabularx}{0.7\linewidth}
        {
        p{0.2\textwidth}@{\hspace{-0.5mm}}
        >{\centering\arraybackslash}X
        >{\centering\arraybackslash}X
        >{\centering\arraybackslash}X
        >{\centering\arraybackslash}X
        >{\centering\arraybackslash}X
        }
        \toprule
                           & \textbf{CIRCO} & \textbf{FIQ-D} & \textbf{FIQ-S} & \textbf{FIQ-T} & \textbf{FIQ-A} \\[-0.3ex]
        \cmidrule(lr){2-2} \cmidrule(lr){3-3} \cmidrule(lr){4-4} \cmidrule(lr){5-5} \cmidrule(lr){6-6}
        \textbf{Methods}   & mAP@5          & R@10           & R@10           & R@10           & R@10           \\[-0.2ex]
        \midrule
        w/o Memory         & 23.79          & 21.49          & 30.98          & 29.63          & 27.37          \\
        Random Vector      & 23.72          & 21.17          & 30.96          & 29.57          & 27.23          \\
        \ccol SRAIN (Ours) & \ccol 26.16    & \ccol 23.22    & \ccol 33.14    & \ccol 31.91    & \ccol 29.42    \\
        \bottomrule
    \end{tabularx}
    \vspace{-2ex}
\end{table}

\noindent\textbf{Robustness of memory bank under domain shift.}
To verify that the memory bank provides useful conditioning even under domain shift in zero-shot CoIR, we compare SRAIN against a variant without the memory bank and one using random vectors as conditioning tokens in~\cref{tab:memory_bank}.
On CIRCO and FashionIQ, SRAIN consistently outperformed the w/o Memory variant and the random vector variant.
These results suggest that the memory bank provides meaningful context even when the target domain differs from the training domain, consistent with prior findings that memory-based prototypes~\cite{sharma2021instance}, pairwise affinity~\cite{zhang2022divide}, and relative similarity~\cite{klabunde2024wild} remain reliable under distribution shift.

\begin{wraptable}{r}{0.49\textwidth}
    \centering
    \vspace{-5ex}
    \caption{Performance on WebVid-CoVR.}
    \label{tab:supp_blip}
    \fontsize{8}{9.5}\selectfont
    \begin{tabularx}{0.49\textwidth}
        {
            p{0.4\linewidth}
            >{\centering\arraybackslash}X%
            >{\centering\arraybackslash}X%
            >{\centering\arraybackslash}X
        }
        \toprule
        \textbf{Methods}  & R@1         & R@5         & R@10        \\ [-0.2ex]\midrule
        CoVR              & 55.95       & 81.22       & 89.05       \\
        \ccol Ours (BLIP) & \ccol 56.91 & \ccol 81.53 & \ccol 88.90 \\[-0.2ex]
        \bottomrule
    \end{tabularx}
    \vspace{-3ex}
\end{wraptable}

\noindent\textbf{Other backbone.}
We also validated SRAIN with the BLIP~\cite{li2022blip} backbone on WebVid-CoVR.
As shown in~\cref{tab:supp_blip}, SRAIN achieved 56.91\% R@1, outperforming CoVR~\cite{webvid2024} with the same backbone.
The component-wise ablation with BLIP, provided in the supplement, shows consistent improvements from each proposed component, confirming that our framework generalizes across different vision-language backbones.

\section{Conclusion}
\label{sec:conclusion}

We presented SRAIN, a novel framework for efficient multimodal fusion in composed image and video retrieval.
To this end, we proposed a sample-wise rank-aware interpolation approach that predicts instance-specific weights for fusing reference and modification embeddings.
SRAIN introduces a rank-aware weight estimation strategy based on ranking behavior within mini-batches and a conditioned weight prediction mechanism that references hard negatives to ensure train-test consistency.
The two-stage training strategy prevents unstable optimization by separating encoder fine-tuning from weight predictor training.
The method achieved state-of-the-art performance on WebVid-CoVR and competitive results on CIRCO and FashionIQ, while significantly reducing query-time latency.
Our work demonstrates that lightweight, sample-wise linear interpolation can achieve superior retrieval performance without relying on computationally expensive architectures.

\noindent\textbf{Acknowledgments.}
This work was supported by
AI Center, Samsung Electronics Co., Ltd.,
and the IITP grants (%
RS-2022-II220290,
RS-2022-II220926,\linebreak
RS-2024-00509258, RS-2024-00469482,
RS-2026-25518317,
RS-2019-II191906%
)\linebreak funded by Ministry of Science and ICT, Korea.

\bibliographystyle{splncs04}
\bibliography{main}
\end{document}